\pdfoutput=1

\documentclass[11pt]{article}

\usepackage[final]{acl}

\usepackage{times}
\usepackage{latexsym}

\usepackage[T1]{fontenc}

\usepackage[utf8]{inputenc}

\usepackage{microtype}

\usepackage{inconsolata}

\usepackage{graphicx}
\usepackage{subcaption}
\usepackage{booktabs}
\usepackage{hyperref}
\usepackage{multirow}
\usepackage{amssymb} 
\usepackage{amsmath} 

\title{Limited Generalizability in Argument Mining:\\State-Of-The-Art Models Learn Datasets, Not Arguments}
\author{Marc Feger\\
  Heinrich-Heine-University \\
  D\"usseldorf, Germany\\
  \texttt{marc.feger@hhu.de}\\\And
  Katarina Boland\\
  Heinrich-Heine-University \\ D\"usseldorf, Germany\\
  \texttt{katarina.boland@hhu.de}\\\AND
  Stefan Dietze\\
  GESIS - Leibniz Institute for the \\ 
  Social Sciences \& Heinrich-Heine-University \\ D\"usseldorf, Germany\\
  \texttt{stefan.dietze@gesis.org}}

\newcommand{\acqua}{ACQUA}
\newcommand{\webis}{WEBIS}
\newcommand{\abstrct}{ABSTRCT}
\newcommand{\arguminsci}{ARGUMINSCI}
\newcommand{\ce}{CE}
\newcommand{\cmv}{CMV}
\newcommand{\finarg}{FINARG}
\newcommand{\iam}{IAM}
\newcommand{\pe}{PE}
\newcommand{\sciark}{SCIARK}
\newcommand{\uselec}{USELEC}
\newcommand{\vacc}{VACC}
\newcommand{\wtp}{WTP}
\newcommand{\afs}{AFS}
\newcommand{\ukp}{UKP}
\newcommand{\aec}{AEC}
\newcommand{\taco}{TACO}

\begin{document}
\maketitle
\begin{abstract}
Identifying arguments is a necessary prerequisite for various tasks in automated discourse analysis, particularly within contexts such as political debates, online discussions, and scientific reasoning. In addition to theoretical advances in understanding the constitution of arguments, a significant body of research has emerged around practical argument mining, supported by a growing number of publicly available datasets. On these benchmarks, BERT-like transformers have consistently performed best, reinforcing the belief that such models are broadly applicable across diverse contexts of debate. This study offers the first large-scale re-evaluation of such state-of-the-art models, with a specific focus on their ability to generalize in identifying arguments. We evaluate four transformers, three standard and one enhanced with contrastive pre-training for better generalization, on 17 English sentence-level datasets as most relevant to the task. Our findings show that, to varying degrees, these models tend to rely on lexical shortcuts tied to content words, suggesting that apparent progress may often be driven by dataset-specific cues rather than true task alignment. While the models achieve strong results on familiar benchmarks, their performance drops markedly when applied to unseen datasets. Nonetheless, incorporating both task-specific pre-training and joint benchmark training proves effective in enhancing both robustness and generalization.
\end{abstract}

\section{Introduction}
Undeniably, discourse gives people the opportunity to express and discuss their beliefs on any topic.

Argument mining, in this sense, is the automatic identification of the structure of inference and reasoning expressed as arguments presented in natural language~\cite{lawrence-reed-2019-argument}.

Although there is no one-size-fits-all answer to \emph{What is an argument?}~\cite{stab-etal-2018-cross}, the idea suggests itself that arguments are latent yet observable and revolve around \emph{how} they are constituted in terms of their logical scaffolding of argument discourse units, rather than \emph{what} specific subject they address. In practice, these elements, whether sentences or sub-sentence segments, are pragmatically assigned functional roles, most commonly claims and premises, and form the fundamental building blocks of an argument~\cite{stab-gurevych-2014-annotating, daxenberger-etal-2017-essence, lawrence-reed-2019-argument, lopes-cardoso-2023-argumentation}.

Consider the example \emph{X should Y, because Z}, such as \emph{Students should study, because it improves grades} or \emph{We should reduce plastic use, because it minimizes ocean pollution}, which illustrates that the manifestation of an argument should ideally rely on structural components conveyed through functional patterns, while remaining agnostic of certain topics or other content-specific elements.

For this reason, one might assert that argument mining, in theory, is applicable across different corpora if the structural signals defining arguments are reliably identifiable from appropriately labeled data. Conversely, in practice, any inability to apply these signals to diverse datasets may expose systematic biases in the field, an issue that has long been informally discussed over coffee breaks.

Generalizability, in this regard, takes high priority, especially at leading NLP conferences such as ACL 2025, as it allows models to make reliable and reasonable predictions on data that does not correspond to their training data. This is especially true for real-world models, which should mimic human-like generalization abilities, where emerging evidence indicates that such models are often fine-tuned to the specifics of established benchmark datasets, leading to unfounded optimism about their improvements \cite{saphra-etal-2024-first}.

Consequently, concerns about vulnerability to shortcut learning \cite{geirhos-2020-shortcut-learning} highlight the broader challenge of evaluating baselines beyond isolated benchmarks \cite{rendle-2019-baselines}.

Argument mining is one such area of natural language processing applications in which the ability to generalize is key. Hence, we ask for:

\begin{itemize}
\item [\textbf{Q1}:] How comparable are the existing benchmark datasets for argument mining?
\item [\textbf{Q2}:] Do state-of-the-art argument mining models generalize to out-of-distribution data from other benchmarks?
\item [\textbf{Q3}:] Do these models acquire a generalizable concept of arguments?
\end{itemize}

In this context, there has been speculation that BERT~\cite{devlin-etal-2019-bert}, known to pay great attention to basic syntax, nouns, and co-references~\cite{clark-etal-2019-bert}, is prone to learning shortcuts when mining arguments~\cite{geirhos-2020-shortcut-learning}, where its generalization is limited to within-topic signals in datasets sharing similar argument and topic structures \cite{thorn-jakobsen-etal-2021-spurious}.

Our aim is not to propose a new formalism for arguments or to pinpoint the best-performing argument mining model, but to use data from previous work in which different theories have been applied to see whether individual efforts and perspectives converge in terms of identifying arguments.

With this being said, we perform the first large-scale experimental assessment of benchmarks, systematically evaluating generalization across diverse argument mining datasets following a comprehensive review of datasets spanning 2008 to 2024.

For our study, we selected BERT~\cite{devlin-etal-2019-bert}, RoBERTa~\cite{liu-2019-roberta}, and DistilBERT~\cite{sanh-2019-distilbert} as exemplary BERT-like models, widely recognized as standard baselines in various areas of natural language processing~\cite{rogers-etal-2020-primer}, including recent research on argument mining~\cite{shnarch-etal-2020-unsupervised, mayer-2020-abstrct, fromm-2021-amsr, alhamzeh-etal-2022-time, feger-dietze-2024-taco}. We also examine WRAP~\cite{feger-dietze-2024-bertweets}, the only transformer whose language representation pre-training is extended by leveraging contrasts of inference and information signals to generalize argument components. Although originally designed for cross-topic generalization on Twitter ($\mathbb{X}$), WRAP does not rely on tweet- or topic-specific features to enhance its generalizability, distinguishing it from the others and making it particularly interesting for research.

In this study, we start by detailing our process of finding argument mining benchmark datasets and explain the selection criteria and justifications in Section \ref{sec:am-benchmark-datasets}. The core characteristics of these datasets, addressing research question \textbf{Q1}, are then examined in Section \ref{sec:categorization}. Next, we describe our experimental setup in Section \ref{sec:experimental-setup}, covering both result generation and the implementation of best practices for significance testing, which form the basis for answering \textbf{Q2 - Q3} in Section \ref{sec:results}. The results of this paper are then discussed in Section \ref{sec:discussion} and concluded in Section \ref{sec:conclusion}.

In order not only to elucidate the process but also to foster discussion that may inspire new approaches for novel datasets and broader generalization of argument mining methods, we contribute:
\begin{itemize}
\item [1.] A survey of argument mining datasets between 2008 and 2024, primarily from the ACL Anthology, that identified 52 relevant papers with datasets from leading NLP conferences.
    \item [2.] The first large-scale re-assessment that combines benchmark evaluations for 17 selected argument mining datasets, including controlled manipulation experiments to determine whether the reported state-of-the-art models (BERT, RoBERTa, DistilBERT, WRAP) actually learn generalizable argument concepts.
    \item [3.] Statistical evidence that shortcut learning undermines generalization in argument mining. Although each of the examined transformers delivers strong results on benchmarks, all struggle to varying degrees when applied to other datasets, with WRAP generally performing slightly better. These challenges are compounded by divergent argument definitions and inconsistent annotations across datasets.
\end{itemize}

\section{Argument Mining Benchmark Datasets}\label{sec:am-benchmark-datasets}
This section outlines the dataset collection and selection process, emphasizing the rationale behind our choice of benchmark datasets for argument mining. The decisions for all 52 datasets reviewed are present in Appendix~\ref{sec:appendix-2}. Additionally, the code and data are available in our repository\footnote{\href{https://github.com/TomatenMarc/Limited-Generalizability}{Limited-Generalizability}\label{ref:repo}}.
\begin{table*}[!htb]
\centering
\resizebox{\textwidth}{!}{%
\begin{tabular}{llllll}
\toprule
Dataset & Paper & Genre & Definition & Arguments & No-Arguments \\
\midrule
ACQUA & \cite{panchenko-etal-2019-categorizing} & Mixed & Argumentative & 1,949 & 5,236 \\
WEBIS & \cite{al-khatib-etal-2016-cross} & Online Debate & Argumentative & 10,804 & 5,543 \\
ABSTRCT & \cite{mayer-2020-transformer-based-am} & Academic & Claim-based & 1,308 & 7,323 \\
ARGUMINSCI & \cite{lauscher-etal-2018-argument} & Academic & Claim-based & 6,554 & 9,548 \\
CE & \cite{rinott-etal-2015-show} & Encyclopedia & Claim-based & 1,546 & 85,417 \\
CMV & \cite{hidey-etal-2017-analyzing} & Online Debate & Claim-based & 979 & 1,593 \\
FINARG & \cite{alhamzeh-etal-2022-time} & Spoken Debate & Claim-based & 4,607 & 8,310 \\
IAM & \cite{cheng-etal-2022-iam} & Mixed & Claim-based & 4,808 & 61,715 \\
PE & \cite{stab-gurevych-2017-parsing} & Academic & Claim-based & 2,093 & 4,958 \\
SCIARK & \cite{fergadis-etal-2021-argumentation} & Academic & Claim-based & 1,191 & 10,503 \\
USELEC & \cite{haddadan-etal-2019-yes} & Spoken Debate & Claim-based & 13,905 & 15,188 \\
VACC & \cite{morante-etal-2020-annotating} & Online Debate & Claim-based & 4,394 & 17,825 \\
WTP & \cite{biran-2011-identification-written-dialoges} & Online Debate & Claim-based & 1,135 & 7,274 \\
AFS & \cite{misra-etal-2016-measuring} & Online Debate & Conclusion-based & 5,150 & 1,036 \\
UKP & \cite{stab-etal-2018-cross} & Mixed & Evidence or Reasoning & 11,126 & 13,978 \\
AEC & \cite{swanson-etal-2015-argument} & Online Debate & Implicit-Markup & 4,001 & 1,374 \\
TACO & \cite{feger-dietze-2024-taco} & Twitter Debate & Inference-Information & 864 & 868 \\
\bottomrule
\end{tabular}
}
\caption{The final 17 datasets that meet the sentential, binary label, and reproducibility criteria, each yielding at least 1,700 instances (850 per label) under a stratified 60/20/20 split, ensuring adequate size for the experiments.}
\label{tab:datasets}
\end{table*}

\subsection{Collection Process}\label{sec:collection-process}
As part of our data collection process, we examined the most recent and relevant survey papers on argument mining, primarily from the ACL Anthology \cite{daxenberger-etal-2017-essence, cabrio-2018-survey, lawrence-reed-2019-argument, vecchi-etal-2021-towards, schaefer-2021-am-twitter-survey, ajjour-etal-2023-topic}, all of which catalog datasets addressing various sub-tasks within the field, where argument identification is a fundamental prerequisite for each.

To expand and back up our dataset collection, we searched Google Scholar and Google Dataset Search for the keyword \emph{argument mining} to find contributions beyond survey papers.

Based on our assessment, we found 52 such papers with datasets, mostly from top NLP conferences like ACL, NAACL, LREC, or EMNLP.

\subsection{Selection Criteria}\label{sec:selection}
The dataset selection process for this paper was conducted in two stages. In the primary inclusion phase, we evaluated all 52 datasets based on:
\begin{itemize}
    \item \textbf{Sentential}: The data and labels are at the sentence-level or aggregatable to this level (e.g., from sub-sentence or token annotations). Tweets were excluded from classical sentence conventions due to their unique structure.
    \item \textbf{Binary}: The dataset assigns binary labels to distinguish argument from no-argument sentences (e.g., based on the presence or absence of claims or other argument components).
    \item \textbf{Reproducible}: The dataset is largely replicable, with minor discrepancies from the publication (e.g., updates or duplicate removal affecting size). To ensure reproducibility, we reviewed documentation, labels, guidelines, and tools, and attempted to resolve access issues (e.g., client-sided or coding errors).
\end{itemize}

We applied these criteria sequentially, excluding datasets immediately upon failing any condition, eliminating 24 of the initial 52. In the refined inclusion step, we assessed relationships and data sufficiency to ensure adequate evaluation and generalization sizes, leading us to consider:

\begin{itemize}
    \item \textbf{Related}: Connections between datasets such as updated versions, additional non-task-related features (e.g., stance added to a claim), and curated subsets derived from repositories that serve as data sources rather than datasets.
    \item \textbf{Sufficiency}: For a stratified 60/20/20 split, each dataset must have at least 500 training instances and 150 evaluation instances per label. An initial analysis revealed that two in five datasets fell short of this threshold, and alternative splits (e.g., 70/15/15 or 80/10/10) would further reduce evaluation sizes, worsening the small-data issue.
\end{itemize}

In total, this process resulted in 17 datasets encompassing \textasciitilde 345k labeled sentences, each meeting the aforementioned criteria. The final selection of datasets included in this study is listed in Table \ref{tab:datasets}.

\section{Characterizing Argument Mining Benchmark Datasets and Definitions}\label{sec:categorization}
Before addressing \textbf{Q1}, we briefly introduce the individual datasets, organizing them by their primary labels. We then give the answer to \textbf{Q1} in terms of comparing definitions in Section \ref{sec:definitions} and textual characteristics in Section \ref{sec:characteristics}.

\textbf{Argumentative} serves as an umbrella term, identifying arguments with markers or patterns that suggest structural components, without necessarily specifying their roles (e.g., as claim or inference). In this sense, \acqua~\cite{panchenko-etal-2019-categorizing} contains 7,185 argumentative sentences from Common Crawl~\cite{panchenko-etal-2018-building}, covering topics like computer science and brands, categorizing comparisons (e.g., Matlab vs. Python) as argumentative or not. Similarly, \webis~\cite{al-khatib-etal-2016-cross} comprises 16,347 segments across 14 topics (e.g., culture, health) from iDebate, with user-assigned labels (introduction, for, against) mapped to argumentative and non-argumentative labels.

\textbf{Claim-based} approaches explicitly annotate for the presence of claims as the core of an argument.  
Thereby, \abstrct~\cite{mayer-2020-transformer-based-am}, sourced from PubMed, comprises 8,631 sentences extracted from abstracts related to five diseases (e.g., neoplasm, glaucoma).  
\arguminsci~\cite{lauscher-etal-2018-argument} provides annotations for the Dr. Inventor dataset~\cite{fisas-etal-2016-multi} for computer graphics publications, totaling 16,102 sentences.  
\ce~\cite{rinott-etal-2015-show} contains 86,963 sentences from Wikipedia across 58 topics (e.g., one-child policy, physical education).  
\cmv~\cite{hidey-etal-2017-analyzing} consists of 2,572 sentences from the \emph{Change My View} subreddit, spanning a diverse range of topics.  
\finarg~\cite{alhamzeh-etal-2022-time} comprises 12,917 sentences sourced from transcribed earnings calls of Amazon, Apple, Microsoft, and Facebook.  
Moreover, \iam~\cite{cheng-etal-2022-iam} contains 66,523 sentences from various online platforms across 123 topics (e.g., vaccination, multiculturalism), while \pe~\cite{stab-gurevych-2017-parsing} includes 7,051 annotated sentences from persuasive essays (e.g., about cloning).  
\sciark~\cite{fergadis-etal-2021-argumentation} contains 11,694 annotated sentences from scientific literature (e.g., PubMed, Semantic Scholar) on sustainable development goals (e.g., well-being, gender equality), also considering generalization to \abstrct.  
On the other hand, \uselec~\cite{haddadan-etal-2019-yes} offers 29,093 sentences from transcripts of U.S. presidential debates from 1960 (Kennedy vs. Nixon) to 2016 (Clinton vs. Trump), transcribed from the Commission on Presidential Debates.  
\vacc~\cite{morante-etal-2020-annotating} offers 22,219 sentences from a mixed collection of online debates about vaccination, while \wtp~\cite{biran-2011-identification-written-dialoges} includes 8,409 sentences from Wikipedia Talk Pages on various topics (e.g., Darwinism, the Catholic Church).

\textbf{Others} represents a residual category encompassing a variety of distinct definitions.  
\afs~\cite{misra-etal-2016-measuring} comprises 6,186 annotated sentences drawn from online debate platforms such as iDebate and ProCon for three topics (e.g., gay marriage, death penalty). Sentences are labeled based on whether they explicitly convey a specific argument facet, with conclusions serving as the core component of the argument.  
\ukp~\cite{stab-etal-2018-cross} contains 25,104 sentences across eight topics (e.g., nuclear energy, minimum wage) for cross-topic argument mining from heterogeneous sources, where arguments provide evidence or reasoning to support or oppose a topic.  
On the other hand, \aec~\cite{swanson-etal-2015-argument} contains 5,375 sentences on four topics (e.g., evolution, gun control) from CreateDebate, highlighting simple argument signals with labels based on the implicit markups: so, if, but, first, I agree that.  
Finally, \taco~\cite{feger-dietze-2024-taco} comprises 1,734 tweets spanning six topics (e.g., abortion, Squid Game). It is designed for cross-topic argument mining on Twitter, focusing on inference to shape arguments.

\subsection{Comparing Argument Definitions}\label{sec:definitions}
\textbf{(Q1)} \emph{Argument definitions vary, reflecting a spectrum of perspectives that contribute to a shared understanding of arguments.}
Central to this is the observation that definitions mutually inform each other in their concepts \cite{lopes-cardoso-2023-argumentation}. For example, in Table \ref{tab:datasets} most papers are claim-based, but when comparing the definitions, some view a claim as argumentative~\cite{lauscher-etal-2018-argument, fergadis-etal-2021-argumentation}, others as conclusive~\cite{mayer-2020-transformer-based-am}, as stances~\cite{rinott-etal-2015-show, hidey-etal-2017-analyzing, cheng-etal-2022-iam, stab-gurevych-2017-parsing}, or as a hybrid concept of all these~\cite{haddadan-etal-2019-yes, morante-etal-2020-annotating}.

Hence, further clarification is needed, especially concerning their generalization as part of \textbf{Q2 - Q3}. Thereby, Table \ref{tab:examples}, with examples from different definitions, illustrates whether their efforts nevertheless converge in the identification of arguments despite different perspectives.

\begin{table*}[!th]
\centering
\renewcommand{\arraystretch}{1.2} 
\resizebox{\textwidth}{!}{%
\begin{tabular}{lll}
\hline
\multicolumn{1}{c}{\textbf{Label}} & \multicolumn{1}{c}{\textbf{Dataset}} & \multicolumn{1}{c}{\textbf{Example}} \\ \hline
\multirow{3}{*}{ARG}               
& \acqua & \begin{tabular}[c]{@{}l@{}}We chose MySQL over PostgreSQL primarily because it scales better and has embedded replication.\end{tabular} \\ \cline{3-3} 
& \sciark & \begin{tabular}[c]{@{}l@{}}In this case, if symptomatic, the treatment should be surgery, clinical follow-up, and counseling.\end{tabular} \\ \cline{3-3} 
& \aec & \begin{tabular}[c]{@{}l@{}}So it would seem that if there is a scientific theory of [\dots], it has been tested [\dots] and therefore [\dots].\end{tabular} \\ \hline
\multirow{3}{*}{$\neg$ARG}              
& \webis & \begin{tabular}[c]{@{}l@{}}The Mo Ibrahim Prize was first established in 2007, and the prize represents [\dots] African leadership.\end{tabular} \\ \cline{3-3} 
& \finarg & \begin{tabular}[c]{@{}l@{}}For those unable to attend in person, these events will be webcast and you can follow [...] at URL.\end{tabular} \\ \cline{3-3} 
& \taco & \begin{tabular}[c]{@{}l@{}}'Bitter truth': EU chief [...] on idea of Brits keeping EU citizenship after \#Brexit URL via USER\end{tabular} \\ \hline
\end{tabular}
}
\caption{Examples of argument (ARG) and no-argument ($\neg$ARG) sentences from various datasets. Despite differences in definitions and topics, the similarities within and distinctions between label groups underscore the shared endeavor of argument mining approaches in identifying arguments, though each emerged differently.}
\label{tab:examples}
\end{table*}

\subsection{Comparing Dataset Dimensions}\label{sec:characteristics}
First, the two text dimensions used to analyze the selected datasets are presented. For dataset-wise correlations of these, please refer to Appendix \ref{sec:appendix-3}.

\textbf{Sentence-Level}: To capture a broad, macro-level view without delving into individual word details, we used spaCy\footnote{\href{https://spacy.io/}{spacy.io}\label{ref:spacy}} to extract key textual attributes. These features reveal the overall structural and statistical properties of sentences, enabling sentence-level characterization of each dataset by:

\begin{itemize}
    \item \emph{Length}: Measured by the number of words per sentence, which serves as an indicator of linguistic complexity and verbosity.
    \item \emph{Stop/Function Word Ratio}: The ratio of stop (e.g., it, is, are) and function words (e.g., against, because, therefore), including discourse markers, to the other words in a sentence to show their relative frequency of use.
    \item \emph{Type-Token Ratio}: The ratio of unique words to total words in a sentence, assessing lexical diversity.
    \item \emph{Readability}: The Flesch Reading Ease score quantifies text clarity, with lower values ($0 \leq$) indicating complex academic language and higher values ($\leq 100$) denoting easy readability, understandable by an 11-year-old.
    \item \emph{Entropy}: Quantifies lexical unpredictability and the amount of information in a sentence, with values ranging from 0 (fully predictable text) to 1 (maximal unpredictability).
    \item \emph{Sentiment}: Defined by polarity, ranging from -1 (extremely negative) to 1 (extremely positive), and subjectivity, ranging from 0 (objective) to 1 (subjective), possibly revealing persuasive strategies through emotions.
    \item \emph{Part-of-Speech Tags}: The distribution of the 17 universal POS tags reflects basic syntax, lexical composition, and stylistic variation.
\end{itemize}

\textbf{Word-Level}: To compare datasets at the word level, we analyze the vocabulary of unique words used in each dataset. We extend this to words that convey the central semantic content of a sentence (e.g., government, abortion, freedom), that is, all words except stop and function words, discourse markers, and punctuation. Their relatedness or uniqueness is described using Jaccard similarity, a measure of similarity between two sets based on the ratio of their intersection to their union.

\textbf{(Q1)} \emph{The sentence structures are strongly correlated across all datasets and labels.}  
On average, a sentence contains 21 words, with nearly every second word (48\%) being a stop or function word. Sentences are lexically diverse (91\% type-token ratio) yet highly readable (63\% readability). The high predictability (22\% entropy) and objective tone (43\% subjectivity) suggest clear, structured writing with a slightly positive inclination (8\% polarity). This is reinforced by the POS patterns, where sentences typically include five nouns, three punctuation marks, and two verbs, adpositions, and determiners, with other tags averaging below two.

Moreover, an average sentence closely aligns with both argument and no-argument sentences across these 24 sentence-level features (Spearman’s $\rho \geq 0.97$), with a strong correlation ($\rho \geq 0.68$) across datasets. Slight differences exist in length, with an argument sentence averaging 24 words compared to 20 for a no-argument sentence, with readability scores of 60\% and 64\%, respectively.

\textbf{(Q1)} \emph{Datasets and labels mainly differ in their semantic content.}
Looking at the vocabularies, the datasets remain largely distinct, with 7–36\% Jaccard similarity, a trend also observed for the semantic content words, reflecting their open-class.

In contrast, stop, function, and discourse words show over 73\% overlap due to their closed nature.

Interestingly, while comparing sentences across labels shows similar patterns, words describing the core semantic content remain largely distinct, overlapping below 48\% and 19\% on average, reinforcing lexical separation.  
Undeniably, the datasets share overlapping content, e.g., when discussing the one-child policy (\pe) and abortion (\iam, \taco, \ukp) or, figuratively speaking, the death penalty (\aec). Similarly, when discussing vaccination (\vacc) overlaps might occur with medical (\abstrct) or sustainability (\sciark) topics.

However, we found that these similarities are not very pronounced and that the datasets and labels are largely disjointed in terms of their core semantic content. This could provide the models with a shortcut opportunity, not based on how the labels are constructed, but rather on what they are about.

\section{Experimental Setup}\label{sec:experimental-setup}
In this section, we outline the experimental setup and the best practices used for statistical testing to generate the data needed to answer \textbf{Q2 - Q3}.

\textbf{Sampling}: To create fixed training, development, and test sets, we used a 60/20/20 stratified split for each of the 17 datasets in Table \ref{tab:datasets}, selecting 850 instances per label, corresponding to 1,700 samples per dataset and 28,900 in total.

\textbf{Transformers}: We selected BERT~\cite{devlin-etal-2019-bert}, RoBERTa~\cite{liu-2019-roberta}, and DistilBERT~\cite{sanh-2019-distilbert} as widely accepted standard baselines for NLP~\cite{rogers-etal-2020-primer}, including argument mining~\cite{shnarch-etal-2020-unsupervised, mayer-2020-abstrct, fromm-2021-amsr, alhamzeh-etal-2022-time, feger-dietze-2024-taco}. Further, we examined WRAP \cite{feger-dietze-2024-bertweets}, the only transformer that is specifically pre-trained for argument generalization. This applies contrastive learning to cluster similar manifestations of inference and information, separate dissimilar ones, and produce generalized embeddings robustly adaptable to downstream classification. However, our goal is to assess the generalizability of these state-of-the-art argument mining models, not to find the best. For these, we use the standard hyperparameter grid for GLUE \cite{wang-etal-2018-glue}, as accepted in the BERT and RoBERTa papers, balancing performance and time with a batch size of 32, 3 epochs, and a learning rate between 2e-5 and 5e-5, each trained on an A100 GPU.

\textbf{Benchmarking and Generalization}: The experiments presented here are the core investigations related to \textbf{Q2}. For each, we report the test results after tuning the hyperparameters to a target's development dataset, optimizing the macro F1 score to ensure equal importance of both labels.

We begin with an initial assessment using pairwise comparisons, following the transfer learning framework \cite{sinno-2010-transfer, houlsby-2019-transnlp, fuzhen-2019-transfer}, where models are trained on one dataset and evaluated on others, including benchmarks on individual datasets. This yields a $17\times17$ matrix per model, with rows as training and columns as test data, see Figure \ref{fig:train-on-one}.

Secondly, we conducted a supplementary experiment by training on all but one dataset and testing on the reserved one, forcing the models to generalize from joint benchmark data \cite{hays-2023-bots, feger-dietze-2024-bertweets}. Thereby, we will report the performance per model and evaluate each against the excluded dataset’s state-of-the-art benchmark, compare Table \ref{tab:delta-leave} and Figure \ref{fig:train-on-one}.

\textbf{Disrupting Argument Signals}: To build on the experiments addressing \textbf{Q2} and provide insight for \textbf{Q3}, we apply controlled input manipulation to both experiments described above. Specifically, we assess transformer performance after systematically removing stop and functional words (e.g., a, the, against, because), discourse markers, and punctuation using spaCy\footref{ref:spacy}. This process results in the elimination of around half the words in each sentence. It is therefore assumed that the removal of these lexical and syntactic elements, which also function as scaffolding for rhetorical and logical devices \cite{knott-1994-relations}, suppresses the linguistic cues that, in theory, enable the distinction between the elements that constitute an argument and those that do not \cite{daxenberger-etal-2017-essence, opitz-frank-2019-dissecting, thorn-jakobsen-etal-2021-spurious}. What remains is a lexical skeleton that primarily reflects topical and subject-related content while omitting functional and discursive elements, calling into question the model's ability to discern argued excerpts from mainly descriptive content \cite{lopes-cardoso-2023-argumentation}, see Table \ref{tab:filtering}.

\begin{table}[!htb]
\centering
\renewcommand{\arraystretch}{1.2} 
\resizebox{\columnwidth}{!}{%
\begin{tabular}{lll}
\hline
\multicolumn{1}{c}{\textbf{Label}} & \multicolumn{1}{c}{\textbf{Form}} & \multicolumn{1}{c}{\textbf{Example}}                                                                                                      \\ \hline
ARG                                & Original                           & \begin{tabular}[c]{@{}l@{}}They should increase more routes to\\make people transport more easily.\end{tabular}                 \\ \cline{3-3} 
                                   & Manipulated                           & increase routes people transport easily                                                                                                               \\ \hline
$\neg$ARG                          & Original                           & \begin{tabular}[c]{@{}l@{}}Should governments spend more money\\on improving roads and highways?\end{tabular} \\ \cline{3-3} 
                                   & Manipulated                           & \begin{tabular}[c]{@{}l@{}}governments spend money improving\\roads highways\end{tabular}                                                                                                                  \\ \hline
\end{tabular}%
}
\caption{Example from \pe\ showing an argument (ARG) and no-argument ($\neg$ARG) sentence in the original and manipulated form.}
\label{tab:filtering}
\end{table}

\textbf{Evaluation}: We perform the experiments for \textbf{Q2 - Q3} and repeat them three times, each with varied samples and training initializations. To test significance, we use a two-way ANOVA with repeated measures for experimental robustness and one-tailed Student's t-tests for pairwise comparisons of models, see Appendix~\ref{sec:statistical-design} for full details.

\section{Results}\label{sec:results}

\begin{figure}[!htb]
    \centering
    \includegraphics[width=\linewidth]{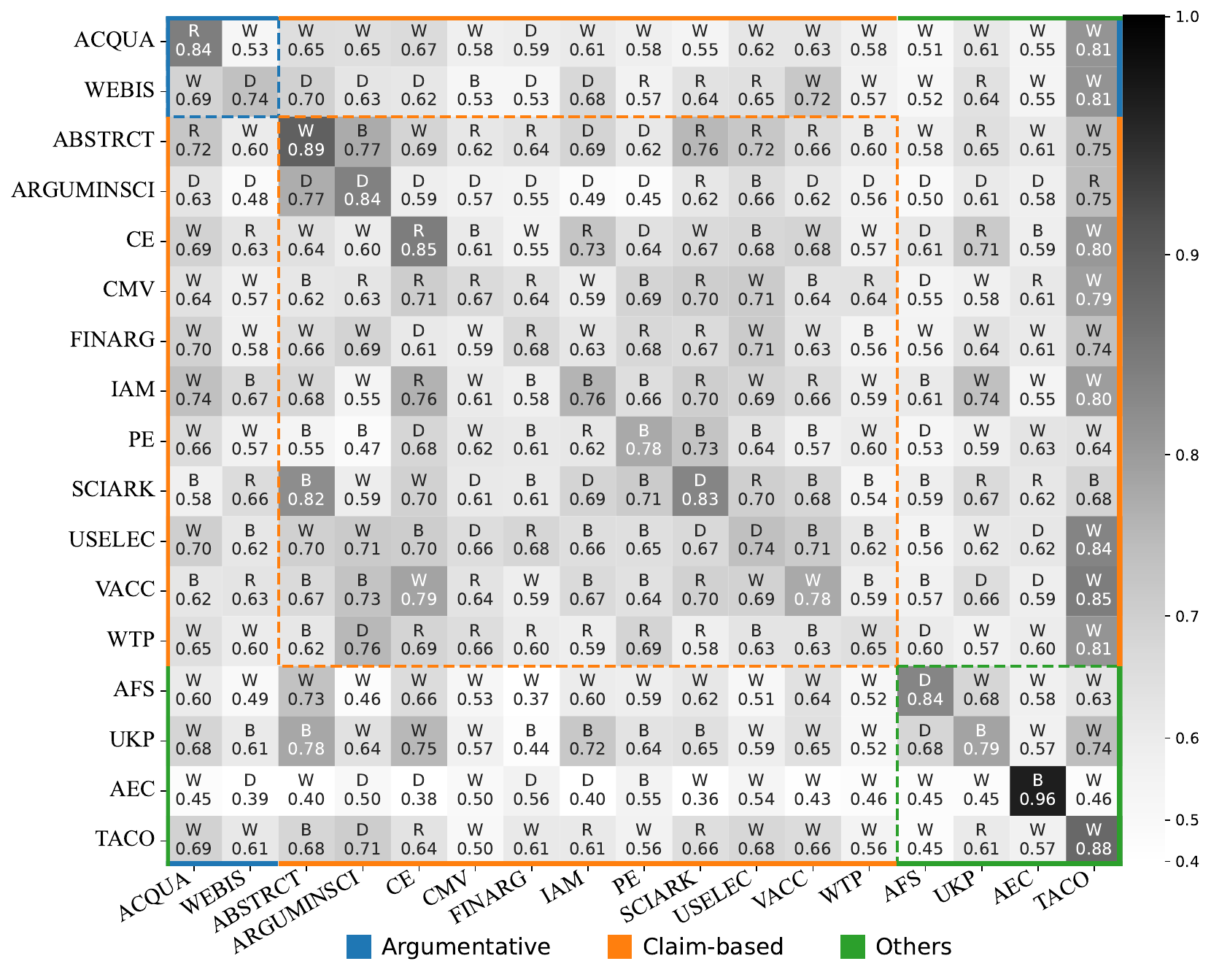}
    \caption{The best macro F1 scores from the benchmarking and pairwise generalization experiments, comparing WRAP (W), BERT (B), RoBERTa (R), and DistilBERT (D), indicate that strong performance is primarily achieved in the benchmark settings, as reflected along the main diagonal. Furthermore, WRAP excels in generalizing to \taco, as seen on the right.}
    \label{fig:train-on-one}
\end{figure}

In this section, we will address and answer questions \textbf{Q2 - Q3}. To this end, we will mainly focus on Figure~\ref{fig:train-on-one}, which compares the pairwise experiments to show which state-of-the-art argument mining model performs best, thus reflecting the current benchmark and generalization landscape. Tying in with this, we will then turn on Table \ref{tab:delta-leave} contrasting the state-of-the-art performance against those obtained by the models if trained on heterogeneous data. In addition, we elaborate on the insights gained from the controlled manipulations applied to these experiments. After that, we will discuss the significance of our results. However, for a better understanding, it can already be assumed that the results for each model and experiment follow a normal distribution, as confirmed with D’Agostino and Pearson’s $K^2$ test ($p \geq .05$).

\textbf{(Q2)} \emph{Strong argument mining baselines do not necessarily imply strong argument generalization.}
A notable observation in Figure \ref{fig:train-on-one} is the contrast between baselines on individual datasets and generalization across multiple datasets and definitions. Strikingly, 97\% of generalization experiments fall below the mean benchmark result ($M = 0.79$), with 62\% scoring under 0.65, while in 8\% of cases generalization drops below 0.5 macro F1, highlighting the challenge of maintaining strong benchmark performances when tested on out-of-distribution datasets. We will further break down our answer:

\textit{Generalizability seems to be the exception rather than the norm.}
Given these circumstances, Table \ref{tab:datasets} shows several notable exceptions of good ($\geq 0.75$) to strong ($\geq 0.8$) generalizability across and within both definitional categories and genres, particularly for claim-based datasets. For instance, strong performance emerges within the academic domain, where \sciark\ reaches 0.82 on \abstrct\ with BERT, and both \abstrct\ and \arguminsci\ achieve 0.77 using BERT and DistilBERT. Evidence of cross-genre generalization also appears in cases such as \iam\ (mixed genre) and \vacc\ (online debate), which achieve 0.76 and 0.79 on \ce\ (encyclopedia) using RoBERTa and WRAP.

Broader generalization across definitions and genres is especially evident in \ukp\ (evidence or reasoning, mixed), which surpasses 0.75 on both \abstrct\ (claim-based, academic) and \ce\ (claim-based, encyclopedia) with BERT and WRAP. Similarly, \taco\ (inference-information, Twitter debate) consistently exceeds 0.8 across a vast range of definitions and genres with WRAP.

Still, both cross-definition and cross-genre generalization remain limited and exceptional.

\textit{Task-related pre-training appears to have a positive effect on overall performance and generalization.}  
Numerically, WRAP ($M = 0.61, SD = 0.1$) shows the best overall performance in terms of macro F1. Notably, WRAP is the only model that attains a mean above 0.6 macro F1, while BERT ($M = 0.58, SD = 0.11$), RoBERTa ($M = 0.57, SD = 0.12$), and DistilBERT ($M = 0.56, SD = 0.11$) all perform worse. This performance advantage is particularly evident in cases where WRAP achieves the highest scores compared to the other models. In fact, WRAP demonstrates superior performance in 133 out of 289 experiments (46\%), whereas BERT does so in 58 experiments (20\%), RoBERTa in 50 experiments (17\%), and DistilBERT in 48 experiments (17\%).

\begin{table}[!htb]
\centering
\renewcommand{\arraystretch}{1.1}
\resizebox{\columnwidth}{!}{%
\begin{tabular}{lllllll}
\toprule
 & WRAP & BERT & RoBERTa & DistilBERT & SOTA & $\Delta_{max/min}$ \\
\midrule
ACQUA & \textbf{0.66} & 0.6 & \textit{0.59} & \textit{0.59} & \textbf{0.84} & 0.18 / 0.25 \\
WEBIS & 0.63 & \textbf{0.66} & \textit{0.62} & 0.65 & \textbf{0.74} & 0.08 / 0.12 \\
ABSTRCT & \textbf{0.74} & \textbf{0.74} & \textbf{0.74} & \textit{0.71} & \textbf{0.89} & 0.15 / 0.18 \\
ARGUMINSCI & \textbf{0.59} & \textit{0.47} & 0.55 & 0.5 & \textbf{0.84} & 0.25 / \underline{0.37} \\
CE & \textbf{0.77} & \textit{0.72} & 0.76 & \textit{0.72} & \textbf{0.85} & 0.08 / 0.13 \\
CMV & \textbf{0.63} & 0.62 & 0.62 & \textit{0.58} & \textbf{0.67} & 0.04 / 0.09 \\
FINARG & \textit{0.61} & 0.62 & \textbf{0.66} & 0.65 & \textbf{0.68} & 0.02 / 0.07 \\
IAM & \textbf{0.73} & \textit{0.71} & \textbf{0.73} & \textbf{0.73} & \textbf{0.76} & 0.03 / 0.05 \\
PE & \textit{0.65} & \textit{0.65} & \textbf{0.69} & \textit{0.65} & \textbf{0.78} & 0.09 / 0.13 \\
SCIARK & \textbf{0.75} & \textit{0.73} & 0.74 & \textit{0.73} & \textbf{0.83} & 0.08 / 0.1 \\
USELEC & \textbf{0.7} & 0.66 & 0.68 & \textit{0.59} & \textbf{0.74} & 0.04 / 0.15 \\
VACC & \textit{0.68} & \textbf{0.7} & \textit{0.68} & 0.69 & \textbf{0.78} & 0.08 / 0.1 \\
WTP & \textbf{0.59} & 0.55 & 0.55 & \textit{0.54} & \textbf{0.65} & 0.06 / 0.11 \\
AFS & \textit{0.57} & 0.58 & 0.59 & \textbf{0.6} & \textbf{0.84} & 0.24 / 0.27 \\
UKP & \textbf{0.7} & \textit{0.67} & \textbf{0.7} & 0.68 & \textbf{0.79} & 0.09 / 0.12 \\
AEC & 0.52 & \textbf{0.57} & \textit{0.51} & 0.56 & \underline{\textbf{0.96}} & \underline{0.39} / \underline{0.45} \\
TACO & \textbf{0.76} & 0.61 & 0.65 & \textit{0.55} & \textbf{0.88} & 0.12 / \underline{0.33} \\
\bottomrule
\end{tabular}
}
\caption{Transformers trained on all but the target benchmark are evaluated against their state-of-the-art baseline (SOTA), compare diagonal of Figure \ref{fig:train-on-one}. \textit{Minimum} and \textbf{Maximum} values indicate deviation from SOTA ($\Delta_{max/min}$). While all models fall short relative to SOTA, WRAP yields the best results in most cases.}
\label{tab:delta-leave}
\end{table}

\textit{Joint benchmark data for training may also help bootstrap reliable and improved generalization.}
Furthermore, the results of the supplementary experiment presented in Table~\ref{tab:delta-leave} indicate that overall performance tends to improve when models are trained on joint benchmark data. Thereby, WRAP ($M = 0.66, SD = 0.07$), RoBERTa ($M = 0.65, SD = 0.07$), BERT ($M = 0.64, SD = 0.07$), and DistilBERT ($M = 0.63, SD = 0.07$) all achieve average macro F1 scores above 0.6, with values that are numerically higher than those observed in the pairwise setup. Again, WRAP shows the most consistent advantage, ranking first in 11 out of 17 experiments (65\%).

\textbf{(Q3)} \emph{State-of-the-art argument mining models are not solely defined by argument signals.}
Following the controlled manipulation in the pairwise setup, all models dropped to similar levels, WRAP and BERT ($M = 0.56$, $SD = 0.09$), DistilBERT ($M = 0.55$, $SD = 0.1$), and RoBERTa ($M = 0.57$, $SD = 0.1$). Similar trends appear post-manipulation in the supplementary experiment for WRAP, RoBERTa, and DistilBERT ($M = 0.62$, $SD = 0.06$), and BERT ($M = 0.61$, $SD = 0.06$). With careful attention to detail:

\textit{Shortcut learning influences generalization of arguments, but task-related pre-training weakens the impact.}  
For the pairwise experiments, BERT and DistilBERT showed almost no changes after manipulating inputs ($\Delta \leq 0.02$), while RoBERTa maintained its performance completely, suggesting that the overall performance of these models is not based on learning how arguments are constituted. In contrast, WRAP, which relies on its task-related pre-training to embed structural argument components across topics, showed the largest drop in macro F1 with $\Delta = 0.05$.

\textit{Jointly integrating benchmark data for training improves generalization and reduces shortcut reliance.}
The impact of WRAP towards robustness of generalization is also true for the supplementary experiment, where WRAP exhibited the largest performance drop ($\Delta = 0.04$) post-manipulation. Nonetheless, RoBERTa and BERT showed similar trends ($\Delta = 0.03$), while DistilBERT showed mostly no changes ($\Delta = 0.01$). Whereas the results in Table \ref{tab:delta-leave} show that each model underperformed relative to the state-of-the-art baselines, a notable pattern still emerged. This is, training on jointly integrated benchmark data raises the average macro F1 score to at least $0.64$ for three out of four transformers and $0.63$ for the lowest-performing model, compared to a maximum of $0.61$ in pairwise transfer, achieved by WRAP. While only WRAP generalizes better in the pairwise setting and is less affected by lexical shortcuts, this advantage persists when trained on joined datasets. However, in this merged setting, RoBERTa and BERT also show improved robustness, despite their stronger reliance on shortcuts in the pairwise setup. Furthermore, average differences remain moderate with $\bar\Delta_{max} = 0.12$ and $\bar\Delta_{min} = 0.18$ while the models learn from heterogeneous data sources.

\textit{Differences in definitions of arguments reinforce the limitations of generalization.}  
However, while signs of shortcut learning are found, it is undeniably not the sole limiting factor. Averaged across all models, misclassification patterns show that arguments are correctly classified 28\% of the time and no-arguments 37\%, suggesting that identifying no-arguments is easier. This is further supported by the lower misclassification rate for no-arguments (13\%) compared to arguments (22\%), highlighting practical differences in argument definitions that affect both generalization and benchmarks (e.g., due to conflicting annotations). This can also be observed when analyzing the misclassifications of individual models. Here, all models misclassify no-arguments as arguments in fewer than 16\% of cases. In contrast, BERT, RoBERTa, and DistilBERT exhibit higher misclassification rates, ranging from 21\% to 26\%, while WRAP misclassifies arguments as no-arguments in 18\% of cases, highlighting its superior generalization ability for arguments.

\textbf{(Q2 - Q3)} \emph{The experiments demonstrate both statistical significance and practical relevance.}  
Repeated experiments support the robustness of these results. Regarding the pairwise experiments, a two-way repeated measures ANOVA for \textbf{Q2} showed a significant effect only when comparing model performances ($F(3,864) = 69.47, \epsilon = 0.56, p_{\text{corr}} < .05, \eta_G^2 = 0.03$), with negligible resampling or interaction effects. 
For \textbf{Q2}, paired one-tailed t-tests also showed that only model comparisons involving WRAP were significant ($p_{\text{corr}} < .05, 8.12 \leq t(288) \leq 10.14$), with moderate effect sizes ($0.39 \leq d \leq 0.49$). Similarly, repeating \textbf{Q3} revealed no significant effects, confirming that once ablated, the models perform comparably overall.
Also, for \textbf{Q3}, when comparing pre- and post-manipulation results per model, only WRAP showed a relevant decrease ($p < .05, t(288) = -8.91, d = -0.49$). In terms of the supplementary experiments, repetition yielded no significant effects pre- and post-manipulation. However, regarding $\mathbf{Q3}$, one-sided paired t-tests revealed significant post-manipulation decreases for WRAP, RoBERTa, and BERT ($p < .05, -5.52 \leq t(16) \leq -2.67, -0.58 \leq d \leq -0.41$), with WRAP showing the strongest effect.

\section{Discussion}\label{sec:discussion}
To summarize the limited generalization in argument mining addressed, Table \ref{tab:delta-baseline} compares the best baseline results pre- and post-manipulation. On average, macro F1 differences remain close, within $\bar\Delta_{max} = 0.07$ and $\bar\Delta_{min} = 0.12$ per model, and in the best cases even exceed benchmark levels.

In the single case of \aec, which relies on only five keywords for arguments, overemphasis on these signals also appears to impair generalization. Although \aec\ attains the highest score (0.96) and experiences the largest post-manipulation drop ($\leq 0.45$,  Table \ref{tab:delta-baseline}), its generalization is limited to 0.63 or even below 0.5, compare Figure \ref{fig:train-on-one}. Given the low performance and minimal differences between pre- and post-manipulation results, BERT, RoBERTa, and DistilBERT do not clearly demonstrate an inherent ability to generalize arguments.

\begin{table}[!htb]
\centering
\renewcommand{\arraystretch}{1.1}
\resizebox{\columnwidth}{!}{%
\begin{tabular}{lllllll}
\toprule
 & WRAP & BERT & RoBERTa & DistilBERT & SOTA & $\Delta_{max/min}$ \\
\midrule
ACQUA & \textit{0.73} & 0.77 & 0.76 & \textbf{0.78} & \textbf{0.84} & 0.06 / 0.11 \\
WEBIS & \textit{0.61} & 0.66 & 0.66 & \textbf{0.67} & \textbf{0.74} & 0.07 / 0.13 \\
ABSTRCT & \textit{0.83} & \textbf{0.87} & 0.84 & \textbf{0.87} & \textbf{0.89} & 0.02 / 0.06 \\
ARGUMINSCI & 0.78 & \textbf{0.79} & \textit{0.77} & \textit{0.77} & \textbf{0.84} & 0.05 / 0.07 \\
CE & \textit{0.75} & 0.79 & 0.77 & \textbf{0.81} & \textbf{0.85} & 0.04 / 0.1 \\
CMV & \textit{0.57} & 0.64 & 0.64 & \textbf{0.65} & \textbf{0.67} & 0.02 / 0.1 \\
FINARG & 0.62 & \textit{0.61} & 0.66 & \textbf{0.69} & \textbf{0.68} & \underline{-0.01} / 0.07 \\
IAM & \textit{0.66} & 0.69 & \textbf{0.71} & 0.7 & \textbf{0.76} & 0.05 / 0.1 \\
PE & \textit{0.66} & 0.67 & 0.71 & \textbf{0.73} & \textbf{0.78} & 0.05 / 0.12 \\
SCIARK & \textit{0.71} & \textbf{0.8} & 0.77 & 0.79 & \textbf{0.83} & 0.03 / 0.12 \\
USELEC & 0.65 & \textbf{0.66} & \textit{0.62} & \textbf{0.66} & \textbf{0.74} & 0.08 / 0.12 \\
VACC & \textit{0.67} & 0.68 & \textbf{0.69} & \textbf{0.69} & \textbf{0.78} & 0.09 / 0.11 \\
WTP & \textbf{0.58} & \textit{0.54} & 0.57 & 0.56 & \textbf{0.65} & 0.07 / 0.11 \\
AFS & \textit{0.78} & \textbf{0.81} & 0.8 & 0.79 & \textbf{0.84} & 0.03 / 0.06 \\
UKP & \textit{0.74} & 0.76 & \textbf{0.78} & \textit{0.74} & \textbf{0.79} & 0.01 / 0.05 \\
AEC & \textit{0.51} & 0.55 & 0.58 & \textbf{0.59} & \underline{\textbf{0.96}} & \underline{0.37} / \underline{0.45} \\
TACO & \textbf{0.77} & \textit{0.76} & \textit{0.76} & \textbf{0.77} & \textbf{0.88} & 0.11 / 0.12 \\
\bottomrule
\end{tabular}
}
\caption{Post-manipulation performance of each transformer compared to state-of-the-art (SOTA) results for baseline experiments per dataset. \textit{Minimum} and \textbf{Maximum} values are highlighted, with $\Delta_{\text{max/min}}$ indicating their deviation from SOTA.}
\label{tab:delta-baseline}
\end{table}

Although these challenges may be widespread, positive examples highlight the potential for future progress. This is particularly evident in cases involving diverse sources and topics (\vacc, \ce, \taco, \ukp, \iam), where \ukp, \iam, and \taco\ already aim for generalizable annotations.

Despite limitations, the need for a unified structural approach to argument analysis becomes apparent. This is reinforced by the effectiveness of methodologies tailored to argument mining, as seen in WRAP’s strong performance, averaging 0.75 when generalizing to TACO from all other datasets (Figure \ref{fig:train-on-one}). Training on joint benchmark data further strengthens these abilities also for the standard transformers, even if numerical results fall short of the rarely doubted state-of-the-art (Table \ref{tab:delta-leave}). Benchmarking should therefore build on combined datasets that capture the task’s general demands, as in GLUE \cite{wang-etal-2018-glue} and instruction-tuning benchmarks \cite{ouyang-2022-instruction, zhang-2024-survey}, for which decoder-based argument mining \cite{cabessa-etal-2025-argument} may be of interest.

\section{Conclusion}\label{sec:conclusion}  
We present the first large-scale re-evaluation of argument mining benchmarks through a generalization lens and evaluate whether the reported performance marks true progress. While structural patterns hold, thematic and content differences between labels and datasets favor shortcut learning. BERT, RoBERTa, and DistilBERT often rely on this to inflate benchmarks, while WRAP shows more resilience, likely due to its pre-training for argument generalization. Training on shared benchmark data further reduces shortcut reliance and improves generalization, notably in combination with WRAP. Our results stress the need to integrate different task demands and suggest re-framing argument mining as a joint generalizability task.

\section*{Limitations}
This study did not separate direct from implicit arguments lacking clear structural and lexical cues, including discourse markers, and based on data analysis, assumed such cases are rare. However, this may affect interpretation, as implicit arguments are likely to depend on topical and content cues.

While we mostly used publicly available datasets, some require granted access.

Additionally, when extraction scripts were unavailable, we derived our procedures from both the available documentation and our understanding of the original process. This was particularly relevant for datasets where \texttt{.ann} files only provided annotated sequence boundaries for larger documents stored in \texttt{.txt} or \texttt{.json} formats. In such cases, we used spaCy\footref{ref:spacy} for sentence boundary extraction, which may produce boundaries that differ from the original assumptions. Nevertheless, we confirmed that over 95\% of the extracted sentences ended with proper punctuation and began with a capital letter. We provide an extraction script\footref{ref:repo} that automatically retrieves and processes all datasets considered.

The reproducibility of the experiments may be constrained by factors such as data size, runtime, and associated costs, with all experiments in this study running \textasciitilde 126 hours on a costly A100 GPU.

\section*{Acknowledgments}
We sincerely thank the anonymous reviewers for their attentive and constructive feedback, which greatly contributed to improving the paper. Cheers!
\bibliography{anthology,custom}

\appendix
\section{Extended Descriptive and Experimental Details}\label{sec:appendix}
This appendix provides additional data and details omitted from Sections \ref{sec:am-benchmark-datasets} and \ref{sec:categorization}.

\subsection{Section 2}\label{sec:appendix-2}
For Section \ref{sec:am-benchmark-datasets} we present the entire decision-making process for the selection of the benchmark datasets used in this work, which is in Table \ref{tab:survey}.

\subsection{Section 3}\label{sec:appendix-3}
\begin{figure}[!htb]
    \centering
    \includegraphics[width=\linewidth]{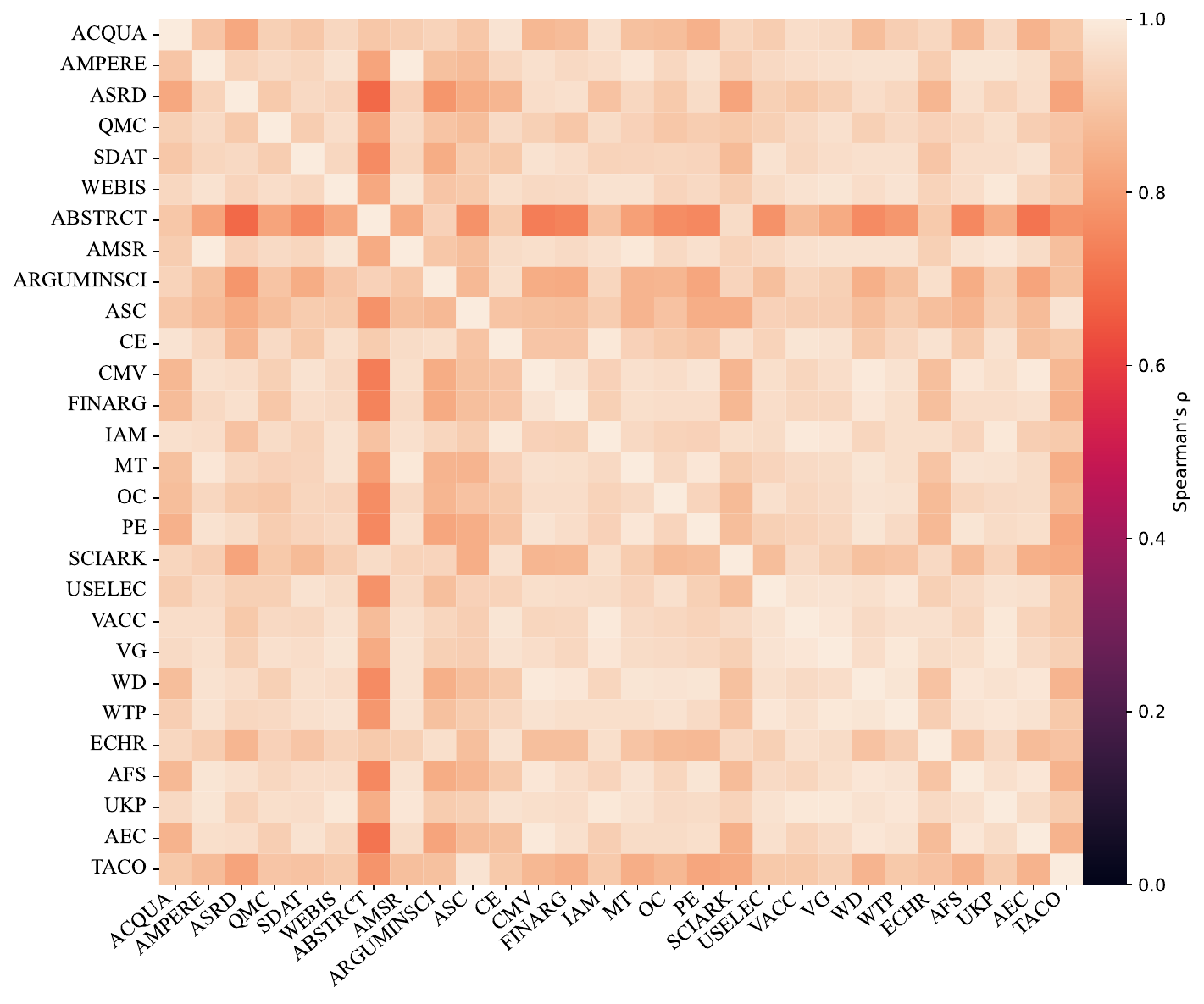}
    \caption{The correlations of the individual datasets (as well as the labels) in relation to the sentence-related features show a strong overall correlation ($\rho \geq 0.68$). Most strikingly, the \abstrct\ dataset stands out as medical texts exhibit different sentence structures from conventional ones, characterized by technical language, methodological details, and numerical values.}
    \label{fig:sentence-corr}
\end{figure}

Figure \ref{fig:sentence-corr} extends the analysis in Section \ref{sec:characteristics} by showing pairwise Spearman’s $\rho$ correlations for all reproducible datasets, including those omitted from experiments due to their small size.

\begin{figure}[!htb]
    \centering
    \includegraphics[width=\linewidth]{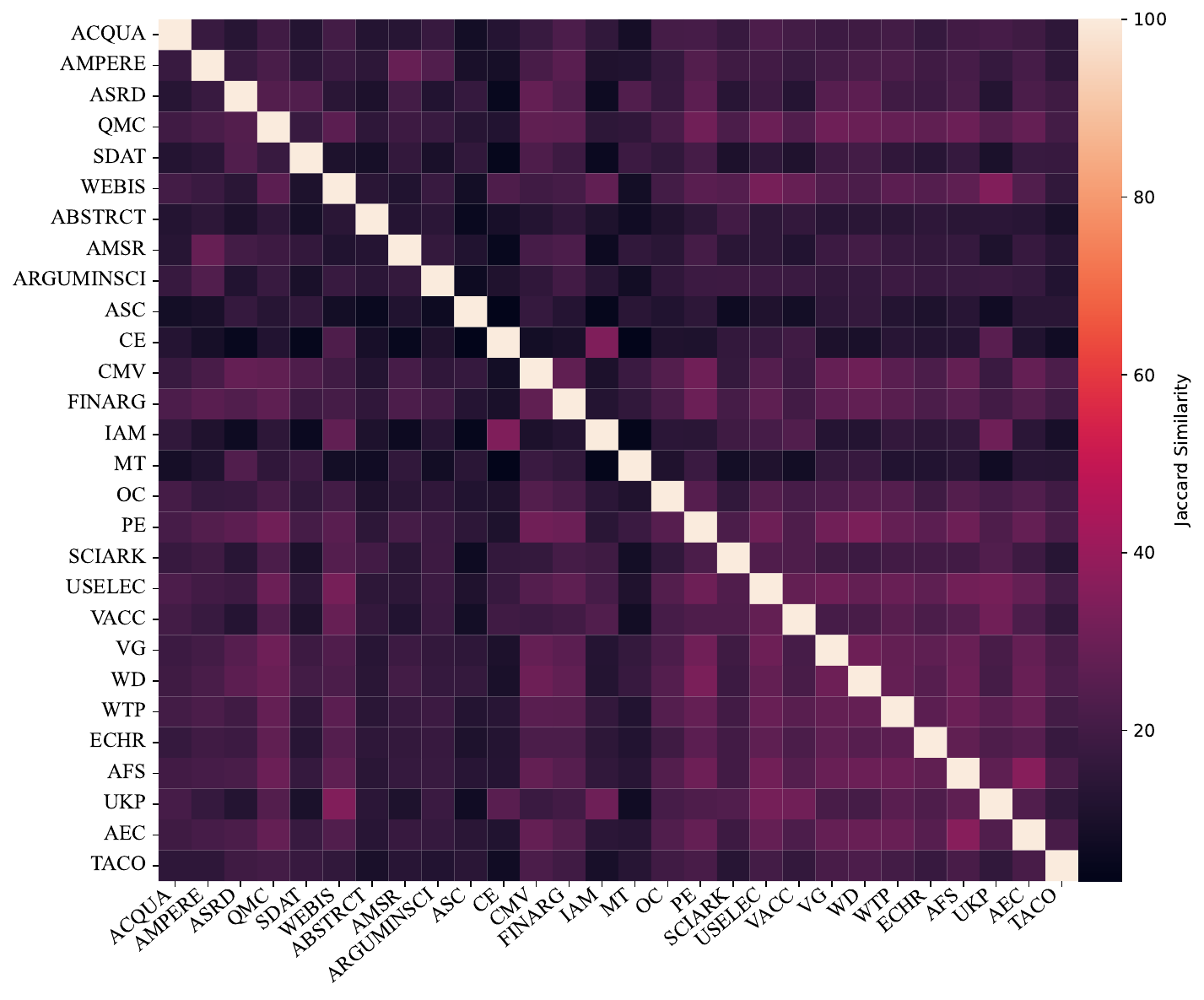}
    \caption{The word overlaps, measured by the Jaccard similarity between the vocabularies of two datasets, show that the datasets (as well as the labels) are generally distinct from each other. The overlaps range between 3–36\%, with an average of 19\%.}
    \label{fig:vocabulary-corr}
\end{figure}

Figure \ref{fig:vocabulary-corr} extends the vocabulary analysis from Section \ref{sec:characteristics} by displaying word overlaps across all datasets with available data.

\section{Statistical Design Protocol}\label{sec:statistical-design}
In this appendix we also explain our protocol for the best-practices of statistical testing as described in Section \ref{sec:experimental-setup} and applied in Section \ref{sec:results}.
\subsection{Two-Way Repeated Measures ANOVA}
We employ a two-way repeated measures ANOVA to evaluate the effects of sampling (factor 1) and model choice (factor 2) on the macro F1 (dependent variable), with each dataset pair treated as a subject.

For valid inference, the following assumptions must be met:
\begin{itemize}
    \item \textbf{Continuous Dependent Variable}: By definition, the macro F1 score is a continuous measure.
    \item \textbf{Within-Subject Design}: Each subject experiences every variation of both factors.
    \item \textbf{Normality}: The dependent variable is approximately normally distributed for each repeated measure (D’Agostino and Pearson’s $K^2$ test).
    \item \textbf{Sphericity}: The variances of the differences between every pair of repeated measures are equal. If the Greenhouse-Geisser $\epsilon$ is below 0.75 (with values near 1 indicating compliance), we adjust the $p$-values ($p_{corr}$).
\end{itemize}

We can specifically evaluate for:
\begin{itemize}
    \item \textbf{Sampling Effect}: Whether variations in data sampling (via different random seeds) influence model performance.
    \item \textbf{Model Choice Effect}: The performance differences among transformer models trained and evaluated on fixed samples. Each model is reinitialized in each trial using distinct random seeds to prevent carry-over effects.
    \item \textbf{Interaction Effect}: Whether the effect of sampling varies across the different models, offering insights into model stability under varying data conditions.
\end{itemize}

We evaluate the practical relevance of statistical significance using the effect size:
\begin{itemize}
   \item \textbf{Generalized Eta Squared ($\eta^2_G$)}: Proportion of the explained variance, interpreted as: \textasciitilde0.01 (small), \textasciitilde0.06 (moderate), \textasciitilde0.14+ (strong).
\end{itemize}

\subsection{One-Tailed Paired Student's t-Tests}  
Further, we conduct one-tailed paired t-tests as post-hoc analysis to identify directional differences (e.g., one model consistently outperforming another). These tests use the same assumptions as the prior ANOVA, except for sphericity. We apply the Bonferroni correction ($p_{\text{corr}}$) for multiple comparisons.  

For these tests, we evaluate their practical relevance using the effect size:  
\begin{itemize}
    \item \textbf{Cohen’s d}: The mean difference between paired conditions relative to the standard deviation of the differences, interpreted as: \textasciitilde0.2 (small), \textasciitilde0.5 (moderate), \textasciitilde0.8+ (strong).
\end{itemize}

\begin{table*}
\centering
\resizebox{\textwidth}{!}{%
\begin{tabular}{lllllllllll}
\toprule
Dataset & Paper & Definition & Genre & Sent. & Binary & Reprod. & Related & Arg. & N-Arg. & Used \\
\midrule
ACQUA & \cite{panchenko-etal-2019-categorizing} & Argumentative & Mixed & Yes & Yes & Yes &  & 1,949 & 5,236 & Yes \\
AMPERE & \cite{hua-etal-2019-argument} & Argumentative & Academic & Yes & Yes & Yes &  & 6,729 & 242 & No \\
ASRD & \cite{shnarch-etal-2020-unsupervised} & Argumentative & Spoken Debate & Yes & Yes & Yes &  & 260 & 440 & No \\
CDCP & \cite{niculae-etal-2017-argument} & Argumentative & Online Debate & Yes & No &  &  &  &  & No \\
COMARG & \cite{boltuzic-snajder-2014-back} & Argumentative & Online Debate & No &  &  &  &  &  & No \\
EDIT & \cite{al-khatib-etal-2016-news} & Argumentative & Online Debate & Yes & No &  &  &  &  & No \\
IAC & \cite{walker-etal-2012-corpus} & Argumentative & Online Debate & No &  &  &  &  &  & No \\
MARG & \cite{mestre-etal-2021-arg} & Argumentative & Spoken Debate & Yes & No &  &  &  &  & No \\
QMC & \cite{levy-etal-2018-towards} & Argumentative & Encyclopedia & Yes & Yes & Yes &  & 733 & 1,766 & No \\
SDAT & \cite{hansen-hershcovich-2022-dataset} & Argumentative & Twitter Debate & Yes & Yes & Yes &  & 387 & 210 & No \\
WEBIS & \cite{al-khatib-etal-2016-cross} & Argumentative & Online Debate & Yes & Yes & Yes &  & 10,804 & 5,543 & Yes \\
AAE & \cite{stab-gurevych-2014-annotating} & Claim-based & Academic & Yes & Yes & Yes & PE &  &  & No \\
ABSTRCT & \cite{mayer-2020-transformer-based-am} & Claim-based & Academic & Yes & Yes & Yes &  & 1,308 & 7,323 & Yes \\
AMECHR & \cite{teruel-etal-2018-increasing} & Claim-based & Legal & Yes & Yes & No &  &  &  & No \\
AMSR & \cite{fromm-2021-argument-mining-peer-reviews} & Claim-based & Academic & Yes & Yes & Yes &  & 839 & 561 & No \\
ARGUMINSCI & \cite{lauscher-etal-2018-argument} & Claim-based & Academic & Yes & Yes & Yes &  & 6,554 & 9,548 & Yes \\
ASC & \cite{wojatzki-2016-tweets} & Claim-based & Twitter Debate & Yes & Yes & Yes &  & 147 & 568 & No \\
CDC & \cite{aharoni-etal-2014-benchmark} & Claim-based & Encyclopedia & Yes & Yes & Yes & CE &  &  & No \\
CE & \cite{rinott-etal-2015-show} & Claim-based & Encyclopedia & Yes & Yes & Yes &  & 1,546 & 85,417 & Yes \\
CMV & \cite{hidey-etal-2017-analyzing} & Claim-based & Online Debate & Yes & Yes & Yes &  & 979 & 1,593 & Yes \\
CS & \cite{bar-haim-etal-2017-stance} & Claim-based & Encyclopedia & Yes & Yes & Yes & CE &  &  & No \\
DT & \cite{olshefski-etal-2020-discussion} & Claim-based & Spoken Debate & No &  &  &  &  &  & No \\
FINARG & \cite{alhamzeh-etal-2022-time} & Claim-based & Spoken Debate & Yes & Yes & Yes &  & 4,607 & 8,310 & Yes \\
IAM & \cite{cheng-etal-2022-iam} & Claim-based & Mixed & Yes & Yes & Yes &  & 4,808 & 61,715 & Yes \\
MT & \cite{peldszus-stede-2015-joint} & Claim-based & Microtext & Yes & Yes & Yes &  & 112 & 337 & No \\
OC & \cite{biran-2011-identification-written-dialoges} & Claim-based & Online Debate & Yes & Yes & Yes &  & 702 & 7,824 & No \\
PE & \cite{stab-gurevych-2017-parsing} & Claim-based & Academic & Yes & Yes & Yes &  & 2,093 & 4,958 & Yes \\
QT & \cite{hautli-janisz-etal-2022-qt30} & Claim-based & Spoken Debate & Yes & No &  & AIFDB &  &  & No \\
RCT & \cite{mayer-2018-argument-rct} & Claim-based & Academic & Yes & Yes & Yes & ABSTRCT &  &  & No \\
SCIARK & \cite{fergadis-etal-2021-argumentation} & Claim-based & Academic & Yes & Yes & Yes &  & 1,191 & 10,503 & Yes \\
UGWD & \cite{habernal-gurevych-2017-argumentation} & Claim-based & Online Debate & Yes & Yes & Yes & WD &  &  & No \\
USELEC & \cite{haddadan-etal-2019-yes} & Claim-based & Spoken Debate & Yes & Yes & Yes &  & 13,905 & 15,188 & Yes \\
VACC & \cite{morante-etal-2020-annotating} & Claim-based & Online Debate & Yes & Yes & Yes &  & 4,394 & 17,825 & Yes \\
VG & \cite{reed-etal-2008-language} & Claim-based & Mixed & Yes & Yes & Yes & AIFDB & 547 & 2,029 & No \\
WD & \cite{habernal-gurevych-2015-exploiting} & Claim-based & Online Debate & Yes & Yes & Yes &  & 211 & 3,661 & No \\
WTP & \cite{biran-2011-identification-written-dialoges} & Claim-based & Online Debate & Yes & Yes & Yes &  & 1,135 & 7,274 & Yes \\
ECHR & \cite{poudyal-etal-2020-echr} & Conclusion-based & Legal & Yes & Yes & Yes &  & 414 & 10,264 & No \\
AFS & \cite{misra-etal-2016-measuring} & Conclusion-based & Online Debate & Yes & Yes & Yes & IAC & 5,150 & 1,036 & Yes \\
ARGSME & \cite{ajjour-2019-argsme} & Conclusion-based & Online Debate & Yes & No &  &  &  &  & No \\
BASN & \cite{kondo-etal-2021-bayesian} & Conclusion-based & Mixed & Yes & No &  &  &  &  & No \\
BIOARG & \cite{green-2018-proposed} & Conclusion-based & Academic & Yes & No &  &  &  &  & No \\
DEMOSTHENES & \cite{grundler-etal-2022-detecting} & Conclusion-based & Legal & Yes & Yes & No &  &  &  & No \\
RSA & \cite{houngbo-mercer-2014-automated} & Conclusion-based & Academic & Yes & No &  &  &  &  & No \\
AIFDB & \cite{lawrence-2012-aifdb} & AIF & Mixed & Yes & No &  &  &  &  & No \\
LAMECHR & \cite{habernal-2023-court-decisions} & Custom Framework & Legal & Yes & No &  &  &  &  & No \\
ABAM & \cite{trautmann-2020-aspect} & Evidence or Reasoning & Mixed & Yes & No &  & AURC &  &  & No \\
ASPECT & \cite{reimers-etal-2019-classification} & Evidence or Reasoning & Mixed & Yes & No &  & UKP &  &  & No \\
AURC & \cite{trautmann-2020-fine-grained} & Evidence or Reasoning & Mixed & Yes & Yes & No &  &  &  & No \\
BWS & \cite{thakur-etal-2021-augmented} & Evidence or Reasoning & Mixed & Yes & No &  & UKP &  &  & No \\
UKP & \cite{stab-etal-2018-cross} & Evidence or Reasoning & Mixed & Yes & Yes & Yes &  & 11,126 & 13,978 & Yes \\
AEC & \cite{swanson-etal-2015-argument} & Implicit-Markup & Online Debate & Yes & Yes & Yes & IAC & 4,001 & 1,374 & Yes \\
TACO & \cite{feger-dietze-2024-taco} & Inference-Information & Twitter Debate & Yes & Yes & Yes &  & 864 & 868 & Yes \\
\bottomrule
\end{tabular}
}
\caption{Summary of the 52 datasets from the reviewed papers, sorted by their applied definitions. Data collection followed the methodology described in Section \ref{sec:collection-process}, and selection criteria are detailed in Section \ref{sec:selection}. Empty entries indicate that the corresponding criteria were not further evaluated because a preceding criterion had already been rejected. The \emph{Related} column indicates connections between datasets, like updates (e.g., AAE to PE, CDC to \ce, RCT to \abstrct), additions of non-task-related features (e.g., CS adds stances to the claims from \ce, ABAM adds aspects to the claims of AURC), or subsets from larger repositories (e.g., VG and QT from AIFDB, \aec\ and \afs\ from IAC).}
\label{tab:survey}
\end{table*}

\end{document}